\documentclass[10pt,twocolumn,letterpaper]{article}

\usepackage[pagenumbers]{cvpr}

\usepackage{cuted}

\usepackage{caption}
\usepackage{multirow}

\usepackage{microtype}

\makeatletter\AtBeginDocument{\renewcommand\paragraph{\@startsection{paragraph}{4}{\z@}{1ex \@plus.3ex \@minus.2ex}{-1em}%
  {\normalfont\normalsize\bfseries}}}\makeatother

\usepackage{xspace}
\newcommand{\best}[1]{\textbf{#1}}
\newcommand{\method}{\textup{Cairn}\xspace}
\newcommand{\pizero}{$\pi_{0.5}$\xspace}
\usepackage{pifont}
\newcommand{\cmark}{\ding{51}}
\newcommand{\xmark}{\ding{55}}

\usepackage{tikz}
\usetikzlibrary{arrows.meta,positioning,fit,calc,backgrounds}

\newcommand{\heldout}{LIBERO-Shift\xspace}
\newcommand{\obs}{\mathbf{o}}
\newcommand{\act}{\mathbf{a}}
\newcommand{\acthat}{\hat{\mathbf{a}}}
\newcommand{\Phimap}{\Phi}
\newcommand{\Tee}{T^{\mathrm{ee}}}
\newcommand{\SE}{\mathrm{SE}(3)}
\newcommand{\SO}{\mathrm{SO}(3)}
\newcommand{\pplan}{p_{\theta}}
\newcommand{\fplan}{f_{\theta}}
\newcommand{\cond}{\mathbf{c}}
\newcommand{\scene}{\mathcal{S}}
\newcommand{\objcloud}{\mathcal{G}}
\newcommand{\startpose}{A}
\newcommand{\instr}{\ell}
\newcommand{\plan}{\boldsymbol{\xi}}
\newcommand{\planstar}{\boldsymbol{\xi}^{\star}}
\newcommand{\wpt}[1]{\mathbf{w}_{#1}}

\definecolor{cvprblue}{rgb}{0.21,0.49,0.74}
\usepackage[pagebackref,breaklinks,colorlinks,allcolors=cvprblue]{hyperref}

\def\confName{CVPR}
\def\confYear{2027}

\title{Cairn: Object-Centric 3D Trajectory Priors for Steering Frozen VLAs}

\author{Agelos Kratimenos$^{1}$ \qquad Kostas Daniilidis$^{1,2}$ \qquad Evangelos Chatzipantazis$^{1}$\\[4pt]
{\normalsize $^{1}$University of Pennsylvania \qquad $^{2}$Robotics Institute, Athena RC}\\[-20pt]
}

\begin{document}
\maketitle
\begin{strip}
\vspace*{-6mm}
\centering
\includegraphics[width=\linewidth]{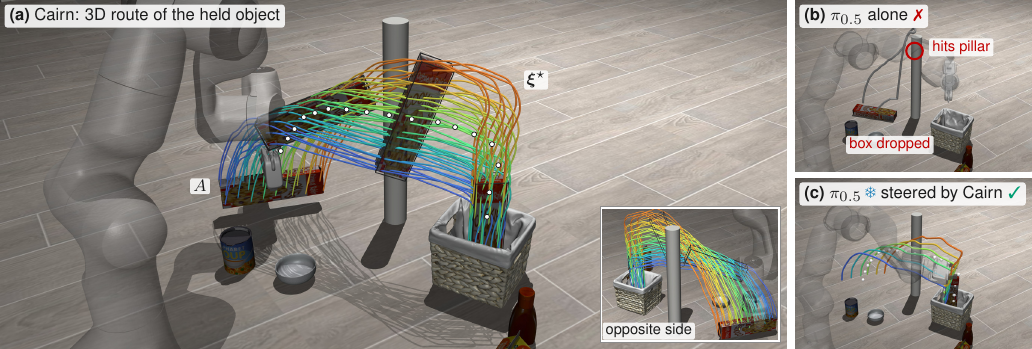}
\captionof{figure}{\textbf{A 3D object route steers a frozen VLA.} (a)~\method's selected $\SE$ route $\planstar$
(dots: waypoints) carries the box from grasp pose $\startpose$ past the pillar; each track follows one box point as the
box turns upright (inset: the other side). (b)~\pizero alone hits the pillar and drops the box. (c)~Steered toward
$\planstar$, frozen \pizero~({\color[HTML]{0072B2}\ding{100}}) places it. Across 1,941 held-out scenes, success
rises from 24.5\% to 95.1\%.}
\label{fig:teaser}
\end{strip}
\begin{abstract}
Vision-language-action models (VLAs) are capable low-level controllers, but they tie where objects go to the layouts
they were trained on: when objects are moved or obstacles block the way, they carry them to the wrong place or into
collisions. Fine-tuning on new layouts needs robot data for each one and erodes skills the model already has. We
introduce \method, which keeps the VLA frozen and steers it at test time along learned 3D trajectories of the object
in its gripper. A small prior reads point clouds of the scene and the held object, along with its pose when grasped and
the instruction, and samples $\SE$ trajectories up to its release that lead it around obstacles and turn it to fit
through gaps. The prior and the VLA form the two levels of a hierarchical planner in which feedback runs both ways: the VLA's intended motion chooses and reshapes the
trajectories, and the chosen one guides the VLA's denoising. Its predicted actions, mapped to the held object's
poses through the grasp, are pulled toward the next waypoint, while the VLA's weights stay untouched. Across 1,941 held-out
LIBERO scenes with displaced objects and obstacles, \method raises the success rate of a frozen \pizero from 24.5\% to
95.1\%, above that of a \pizero fine-tuned on our training layouts, without hurting standard LIBERO. On the
third-party LIBERO-PRO benchmark, it raises success from 27.3\% to 38.0\%, and from 54.4\% to 81.1\% on the episodes it
steers. On a
real Franka, with a prior trained purely in simulation, it guides the frozen DROID \pizero around an obstacle and to a
placement at the limit of the arm's reach, where it otherwise fails. All code, models and data will be made public.
\end{abstract}
\vspace{-4mm}
\section{Introduction}
\label{sec:intro}

Moving into a new home calls on two abilities: lifting and setting down a bookshelf are skills one brings along,
while where it should stand and how to get it there, around the sofa and tilted through the doorway, depend on the new
home. Vision-language-action models
(VLAs)~\cite{brohan2023rt2,kim2024openvla,black2024pi0,physicalintelligence2025pi05,nvidia2025gr00t} have the first:
trained by imitation on large robot datasets, they grasp and place many objects from images and an instruction. They
largely lack the second. Deciding one short action chunk at a time, mostly from 2D images, a VLA absorbs where objects
belong and how to reach them from its training layouts; moving objects collapses its
success~\cite{zhou2025liberopro,fei2025liberoplus,wang2026liberox,emukpere2026brittle}, and obstacles cause
collisions~\cite{song2026omniguide,liu2026vls}. An obstacle calls for a detour chosen before it is
reached, and a long object may need to turn to fit through a gap. Both depend on the scene's 3D geometry and the
object's shape. The
motor skill is in place; the route is missing.

Fine-tuning a VLA can teach it those layouts, but it needs robot demonstrations of every new scene and can make it
forget skills it already has~\cite{zhu2026vlaforgetting}. Guidance instead changes what a generative model produces without
changing the model: each denoising step is nudged along the gradient of an objective. It has spread from image generation~\cite{dhariwal2021guidance} to inverse
problems~\cite{chung2023dps,song2023pigdm}, trajectory planning~\cite{janner2022diffuser} and robot
policies~\cite{wang2025itps,du2025dynaguide,liu2026vls}. Since current VLAs generate actions with flow-matching
experts~\cite{black2024pi0,physicalintelligence2025pi05,nvidia2025gr00t}, the same recipe applies to them with no robot
data or training: OmniGuide~\cite{song2026omniguide} steers frozen \pizero and GR00T by back-propagating 3D energies
through the robot's kinematics. Its targets are local: a collision field does not know the goal, an attractor
pulls straight through obstacles and can get stuck in local minima, and both act on the gripper, not the object it
holds. The open question is not how to steer a frozen VLA, but where to.

A route that helps a frozen VLA must meet several requirements, and existing planners each meet only some of them. It
must be metric and three-dimensional, unlike image paths, which lack
depth~\cite{gu2024rttrajectory,li2025hamster}. It must describe the held object and how it turns, not only the gripper
positions of 3D path predictors~\cite{hwang2026hamster3d,huang2026threedway}. Models of 6-DoF object
motion~\cite{yoshida2025egoscaler,zeng2026flowhoi,saroha2026egoflow} plan the object, but learn from human data, without
its shape or the robot's reach. It must cover the whole carry and find the goal from the instruction, whereas motion
planners~\cite{kuffner2000rrtconnect,sundaralingam2023curobo} must be given the goal. It must reach the VLA without
retraining it, whereas hierarchical VLAs~\cite{li2025hamster,hwang2026hamster3d,bu2024robodual} train their low-level
policy to follow plans. Above all, the route and the VLA must agree. Together they form a hierarchy, a planner of the route above a planner
of actions, and in hierarchies the high level usually commands while the low level is trained to obey. A frozen VLA has its
own preferred motion, and a route chosen without it has to be forced on it. A high level that offers a distribution of
routes can instead couple with the VLA at test time, with feedback in both directions: the VLA's intended motion chooses and reshapes the route, and the
route steers the VLA.

We present \method, a framework that steers frozen VLAs with 3D trajectories of the held object that account for the
scene, its obstacles and the object's own shape. The name comes from the stacked stones that mark a hiking trail: a
cairn shows the way but carries no one along it. In the same way, \method marks the route of the held object and
leaves the moving to the VLA. To meet the requirements above, it builds a two-level planner from a small 3D
specialist, which plans the route, and the frozen 2D generalist, which plans the actions (\cref{fig:overview}). The specialist is a motion prior built like 3D point-cloud
policies~\cite{gervet2023act3d,gkanatsios2025flowmatchactor}, but it predicts where the held object travels rather than
how the robot moves. From point clouds of the scene and the
held object, along with its pose at the grasp and the instruction, it samples with rectified
flow~\cite{liu2023rectifiedflow} a route to the release as $\SE$ waypoints. It is trained without robot actions, on
simulated demonstrations generated with moved objects, obstacles and several routes per scene. At test time the prior and the VLA
steer each other. The VLA's intended motion weights and refines the sampled routes that clear the scene. In turn, at
every denoising step, the object motion that the VLA's action chunk would produce is pulled toward the next waypoint of
the chosen route. The VLA's weights are never changed.

We evaluate \method in simulation and on a real robot. On \heldout, 1,941 held-out LIBERO scenes with moved objects
and obstacles, it raises the success of the frozen \pizero from 24.5\% to 95.1\%, more than fine-tuning
\pizero on the same layouts, which also degrades \pizero on standard LIBERO, where \method leaves performance
unchanged. On LIBERO-PRO~\cite{zhou2025liberopro}, a third-party benchmark of moved objects, it raises \pizero's
success on the episodes it steers from 54.4\% to 81.1\%.
The coupling between the two levels helps where success has room: under LIBERO-PRO's layout shift it adds
2.3 points over a route fixed at the grasp, and where the VLA's intent changes the route it makes motion
smoother and safer. On a real Franka, with a
prior trained only in simulation, \method steers the frozen DROID \pizero around an obstacle and to a
placement at the limit of the arm's reach, where \pizero alone fails. Our contributions are:
\begin{itemize}
  \item \method, to our knowledge the first framework that steers a frozen VLA with learned $\SE$ routes of the object it
    holds, from grasp to release, conditioned on the scene, the object's shape and the instruction.
  \item A hierarchical planner whose two levels, a learned route prior and a frozen VLA, inform each other at test time
    in both directions, without retraining either.
  \item A generator of varied LIBERO demonstrations, with moved objects, obstacles that force detours and turns, and
    several routes per scene, and \heldout, a held-out benchmark built with it.
  \item Extensive experiments in simulation and on a real robot showing that \method makes frozen VLAs robust to moved
    objects and obstacles, outperforms fine-tuning without its forgetting, and carries over to a third-party
    benchmark.
\end{itemize}
\section{Related work}
\label{sec:related}

\paragraph{Vision-language-action models.}
VLAs adapt pre-trained vision-language models to map images and an instruction to robot actions
\cite{brohan2023rt2,kim2024openvla,black2024pi0,physicalintelligence2025pi05,nvidia2025gr00t}. The $\pi$ family
generates action chunks with a flow-matching expert \cite{black2024pi0,physicalintelligence2025pi05}. Near-saturated
LIBERO scores \cite{liu2023libero} hide a reliance on the training layouts: moving objects or changing initial states
collapses success \cite{zhou2025liberopro,fei2025liberoplus,wang2026liberox,emukpere2026brittle}. Fine-tuning on more
varied data, \eg generated procedurally \cite{mandlekar2023mimicgen,garrett2024skillmimicgen,xue2025demogen}, needs
robot actions for every new layout and can cause forgetting \cite{zhu2026vlaforgetting}. Hierarchical VLAs train a
low-level policy to follow a 2D or 3D end-effector path
\cite{gu2024rttrajectory,li2025hamster,lee2025molmoact,hwang2026hamster3d,huang2026threedway,liu2026gae} or a latent plan
\cite{nvidia2025gr00t,zhang2024hirt,bu2024robodual,cercola2026gpl} from a high-level model. The planner can learn from
non-robot data, but the coupling is fixed in training and relearned for every embodiment, executors may ignore the plan
\cite{xie2026fastplans}, and the plan gives the gripper's position, not the held object's pose. Feedback runs only from the plan down, whereas
hierarchical planners also revise the high level with low-level outcomes: route progress~\cite{plaku2010syclop},
predicted control costs~\cite{nair2020hierarchicalforesight}, or low-level proposals during joint diffusion
sampling~\cite{hao2025chd}. DreamTrajectory
\cite{yang2026dreamtrajectory} fine-tunes \pizero to predict a short end-effector trajectory and keeps the chunks
that realise it, so it cannot add scene geometry the policy lacks. We instead couple a frozen generalist VLA at test
time, in both directions, with a learned prior over the held object's $\SE$ route, trained without robot actions.

\paragraph{Test-time steering.}
Diffusion and flow samplers can be biased without retraining by guidance \cite{dhariwal2021guidance} and its variants on
the clean-sample estimate \cite{chung2023dps,song2023pigdm,ye2024tfg,feng2025flowguidance}, inpainting
\cite{lugmayr2022repaint} or products of models \cite{hinton2002poe,liu2022composable,du2023reduce}. Most methods that
steer robot policies add the gradient of an energy to the denoising update and differ in where the target comes from:
human goal points and sketches \cite{wang2025itps}, goal images scored by a learned dynamics model
\cite{du2025dynaguide}, VLM-selected keyframes \cite{hao2025disco} or VLM-written stage rewards \cite{liu2026vls}, a
learned critic \cite{yang2026gaf}, collision costs \cite{li2025lano3dp,guo2026nudge}, or, to hide latency, the previous
action chunk \cite{black2025rtc}. Gradient-free methods rank sampled actions with a value function or a verifier
\cite{nakamoto2024vgps,song2026fmsteer,kwok2025robomonkey,liu2024bid,cui2026dreamsteer,li2025vlapilot}, act on the
initial noise \cite{wagenmaker2025dsrl}, or compose policies \cite{cao2026compose}.
Pretrained generative models have also been coupled at test time, by soft terms on their clean estimates
\cite{alzayer2026coupled}, coupling costs \cite{luan2026pcd}, guidance through a shared embedding
\cite{xing2024seeinghearing} or a known measurement \cite{chung2023blinddps,li2025blindpnpdm}, Gibbs sampling over
projections of a shared state \cite{wang2026ggpa}, or joint sampling of a planner with a model-based
optimiser \cite{jung2025jm2d}. OmniGuide \cite{song2026omniguide} back-propagates 3D energies
through differentiable kinematics into the denoising of frozen \pizero and GR00T: a collision repeller, an attractor at
a VLM-pointed target, and end-effector waypoints from a human video or a motion plan. Its authors note that such fields
cannot reliably avoid local minima. Like OmniGuide, we steer through the robot's kinematics, but toward a global route
of the held object, including its rotation, from a learned prior that is itself steered by the VLA's intent, with no
VLM or human video at test time.

\paragraph{Motion priors for manipulation.}
3D policies \cite{gervet2023act3d,ke2024diffuseractor,gkanatsios2025flowmatchactor,ze2024dp3} predict actions from
calibrated point clouds; they are sample-efficient and precise but trained per benchmark. Our planner adopts their architecture to predict the held object's route instead of actions. Object-centric
methods turn predicted object motion into robot motion: point flow
\cite{yuan2024generalflow,bharadhwaj2024track2act,xu2024im2flow2act,zhi2025flowaction}, $\SE$ object poses from
demonstrations \cite{hsu2025spot} or generated videos \cite{patel2026rigvid,zhang2026emboalign} mapped to the gripper by
a rigid grasp, or poses that condition a trained action head \cite{su2024mba}. Models of 6-DoF object motion, learned
from human egocentric and hand-object data, generate a trajectory from language and the scene
\cite{yoshida2025egoscaler,ding2026dreamtraj,zeng2026flowhoi} or complete an observed one toward a given goal pose
\cite{saroha2026egoflow,zeng2026gmt}. They describe the held object by its pose, category or bounding box rather than
its shape, avoid collisions at most with its centre \cite{saroha2026egoflow}, do not check that a robot can execute the
motion, and when executed are retargeted open loop in place of the controller \cite{zeng2026flowhoi}. Motion planners
compute collision-free robot paths to a given goal by sampling or optimisation
\cite{kuffner2000rrtconnect,sundaralingam2023curobo} or from learned diffusion priors
\cite{janner2022diffuser,carvalho2023mpd,pmlr-v283-chatzipantazis25a}. VLMs can supply the goal as value maps, keypoint
constraints or 3D traces \cite{huang2023voxposer,huang2024rekep,zhou2026robotracer} for a planner to execute. Our prior
learns from robot-executed routes, sees the held object's shape, infers the goal from language, and steers a policy
rather than replacing it.
\section{Method}
\label{sec:method}
\begin{figure*}[t]
\centering
\includegraphics[width=\linewidth]{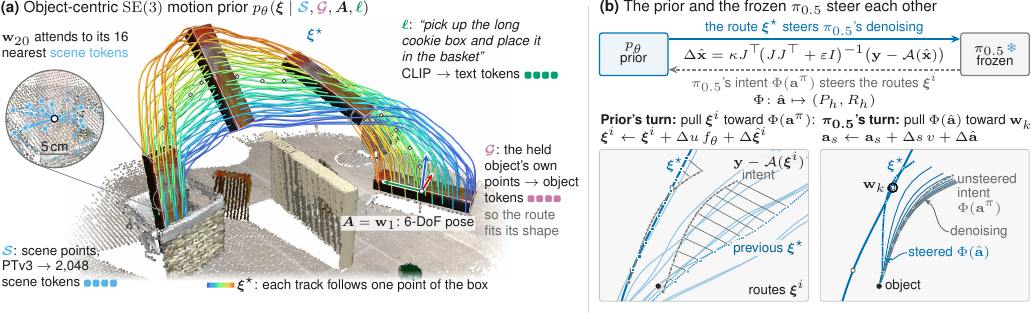}
\vspace{-9mm}
\caption{\textbf{Overview.} \textbf{(a)}~From scene points $\scene$, object points $\objcloud$ at grasp pose $\startpose$
and instruction $\instr$, the prior samples routes of $\SE$ waypoints attending to nearby scene tokens
(\cref{sec:planner}). Each track of the plan $\planstar$ follows one object point. \textbf{(b)}~The prior and frozen
\pizero steer each other (\cref{sec:steering}): \pizero's unsteered intent $\Phimap(\act^{\pi})$ pulls the routes
$\plan^i$ and moves $\planstar$ to its group, and at each denoising step $\Phimap(\acthat)$ is pulled toward
$\wpt{k}$.}
\vspace{-4mm}
\label{fig:overview}
\end{figure*}

We consider rigid pick-and-place: given an instruction, the robot grasps an object and carries it to a placement.
The frozen VLA stays the controller; what it lacks in a new layout is where the object should go. A
learned motion prior proposes routes of the held object (\cref{sec:planner}). It is trained on
generated demonstrations (\cref{sec:data}), and at test time the prior and the VLA steer each other
(\cref{sec:steering}, \cref{fig:overview}).

\subsection{Object-centric SE(3) planning}
\label{sec:planner}

The prior is a distribution over routes of the held object from grasp to placement. Given RGB point clouds of the scene
$\scene$ and of the held object $\objcloud$, the object's pose at the grasp $\startpose=(\mathbf p_A,R_A)$, and the
instruction $\instr$, it samples
\begingroup\setlength{\abovedisplayskip}{4pt plus 1pt minus 1pt}\setlength{\belowdisplayskip}{4pt plus 1pt minus 1pt}%
\begin{equation}
  \plan=(\wpt{1},\ldots,\wpt{K})\sim\pplan(\cdot\mid\cond),\qquad \cond=(\scene,\objcloud,\startpose,\instr),
\label{eq:prior-route}
\end{equation}\endgroup
with $\wpt{k}=(\mathbf p_k,R_k)\in\SE$ and $\wpt{1}=\startpose$. The last pose is the placement, so the goal is inferred
with the route. The prior must (i) describe the held object's motion, including its rotation; (ii) keep
the object clear of the scene; (iii) infer the goal from the instruction and the scene; and (iv) represent
several valid routes rather than their average.

\begin{figure*}[t]
\centering
\includegraphics[width=\textwidth]{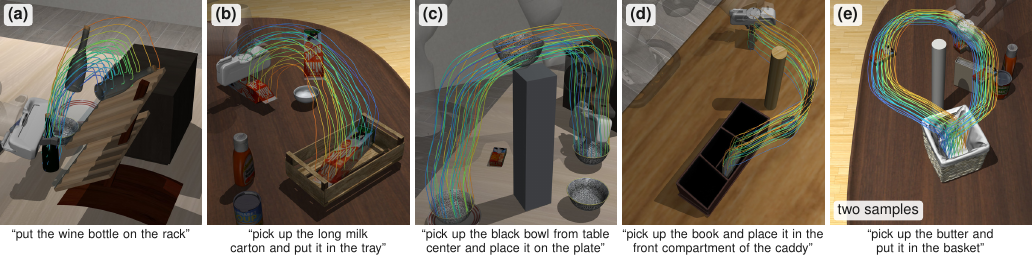}
\vspace{-5.5mm}
\caption{\textbf{Routes proposed by the prior on held-out scenes.} Each panel draws one sampled route as 3D tracks of
points on the held object, with copies of the object along the route, and (e) draws two samples that pass the pole on
opposite sides. Routes were picked for legibility among the 16 samples, and each is clear and ends in the instructed
receptacle.}
\vspace{-2mm}
\label{fig:routes}
\end{figure*}

Each waypoint gives the object's position and orientation, because orientation decides how the object fits the scene:
the box turns upright past the pillar in Fig.~\ref{fig:teaser}(a,c), and the carton rotates into the tray in
Fig.~\ref{fig:routes}(b). Positions are centred on the grasp position $\mathbf p_A$, in the scene and object clouds as
well as the route, with world-aligned axes. Each waypoint is encoded as
$\mathbf y_k=[N_{\rm pos}(\mathbf p_k-\mathbf p_A);\rho(R_k)]\in\mathbb R^9$, where $N_{\rm pos}$ is a fixed
normalisation and $\rho$ the continuous 6D rotation representation~\cite{zhou2019rotation}, and the route as
$\mathbf Y=[\mathbf y_1,\ldots,\mathbf y_K]^\top$. We denote world-frame route decoding by
$\operatorname{dec}_A$, which reverses position normalisation and centring and reconstructs rotations, giving
$\plan=\operatorname{dec}_A(\mathbf Y)$. Waypoints are spaced by an arc length that combines translation and
rotation, so a pure turn also receives waypoints and the route describes progress rather than speed.

The prior builds on 3DFA~\cite{gkanatsios2025flowmatchactor} but reads 3D geometry directly and gives the held object a
branch of its own. The object's cloud is removed from the scene and encoded separately by a point MLP, posed at the
grasp, so the prior sees the shape of what it carries. The scene is encoded from its cloud by a Point Transformer
V3~\cite{wu2024ptv3}, where 3DFA lifts features from camera images. Scene and object tokens attend to each other and
to CLIP features of the instruction~\cite{radford2021clip}, and keep their 3D positions for spatial attention. The start
pose enters as a token that embeds $\startpose$ itself, where 3DFA uses a learned query. At flow time, it modulates
3DFA's flow head in place of the gripper's state. The encoder runs once per input and is reused at every sampling step.

Which nearby geometry matters changes as each waypoint moves during sampling. We augment the inherited global
attention with a local update that gives each waypoint explicit access to the relative scene geometry around it.
Waypoint $k$, with feature $\mathbf h_k$ after global attention and current centred position
$\widetilde{\mathbf p}_k$ at flow time $u$,
attends to its nearest scene tokens $\mathcal N_k$ at positions $\mathbf s_j$. The neighbourhood is recomputed at every
sampling step while the scene features stay cached. The query $\mathbf q_k$ is a projection
of $\mathbf h_k$, and the keys $\mathbf k_j$ and values $\mathbf v_j$ are projections of the scene tokens. Each offset
is embedded as $\mathbf e_{kj}=\mathrm{MLP}((\mathbf s_j-\widetilde{\mathbf p}_k)/r_0)$ and added to both keys and
values:
\begingroup\setlength{\abovedisplayskip}{4pt plus 1pt minus 1pt}\setlength{\belowdisplayskip}{4pt plus 1pt minus 1pt}%
\begin{equation}
\begin{aligned}
  \alpha_{kj}&=\underset{j\in\mathcal N_k}{\operatorname{softmax}}\;\mathbf q_k^\top(\mathbf k_j+\mathbf e_{kj})/\sqrt d,\\
  \mathbf h_k&\leftarrow\mathbf h_k+\mathbf g\odot\mathcal O\Big(\textstyle\sum_{j\in\mathcal N_k}
     \alpha_{kj}(\mathbf v_j+\mathbf e_{kj})\Big).
\end{aligned}
\label{eq:local}
\end{equation}\endgroup
Offsets in the keys decide which neighbours receive attention, and offsets in the values carry the relative geometry
into the waypoint's feature. An output projection $\mathcal O$ and a gate $\mathbf g$, conditioned on the flow time and
the start pose, shape the update. Separate position and rotation branches then predict each waypoint's velocity $\fplan$.

We train the velocity field with rectified flow~\cite{liu2023rectifiedflow,lipman2023flowmatching}. For a training
route $\mathbf Y$ with condition $\cond$, noise $\boldsymbol\epsilon\sim\mathcal N(0,I)$ and flow time $u\in[0,1]$,
every waypoint after the first is interpolated as $\mathbf Y_u=(1-u)\boldsymbol\epsilon+u\mathbf Y$, while the first
stays at the start pose. The field learns the velocity $\mathbf Y-\boldsymbol\epsilon$ on these waypoints. An error in
the 6D coordinates is not an error in orientation, so a geodesic term on the rotations $\widehat R_k$ of the
clean-route estimate $\widehat{\mathbf Y}_u=\mathbf Y_u+(1-u)\fplan(\mathbf Y_u,u,\cond)$ adds orientation supervision
near $u=1$, where that estimate is reliable:
\begingroup\setlength{\abovedisplayskip}{4pt plus 1pt minus 1pt}\setlength{\belowdisplayskip}{4pt plus 1pt minus 1pt}%
\begin{equation}
\begin{aligned}
  \mathcal L&=\mathbb E\,\big\|\fplan(\mathbf Y_u,u,\cond)-(\mathbf Y-\boldsymbol\epsilon)\big\|^2\\
  &\quad+\lambda_R\,\mathbb E\Big[\mathbf 1_{u>u_R}\tfrac{1}{K-1}\textstyle\sum_{k\ge2}d_{\SO}(\widehat R_k,R_k)\Big],
\end{aligned}
\label{eq:prior-fm}
\end{equation}\endgroup
where $d_{\SO}$ is the angle of the relative rotation. At test
time, $M$ routes are sampled by integrating $d\mathbf Y_u/du=\fplan(\mathbf Y_u,u,\cond)$ from Gaussian noise with a
few Euler steps, resetting the first waypoint after each step.

\subsection{Generated training data}
\label{sec:data}

The prior can learn goal-following, obstacle-aware routes only if its data contains them.
LIBERO's~\cite{liu2023libero} do not: the roughly 50 human demonstrations of a task share one layout and no obstacles,
and their routes differ by only 2--4\,cm on average, so they teach one route per task.

\begin{figure}[t]
\centering
\includegraphics[width=\columnwidth]{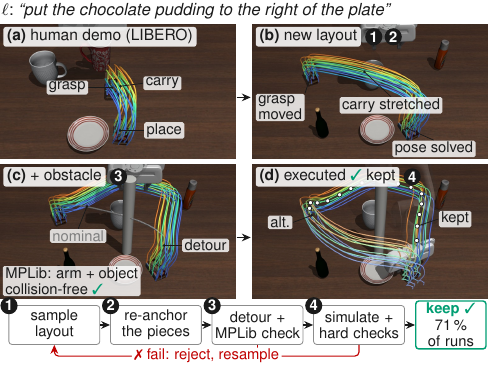}
\vspace{-2em}
\caption{\textbf{Automatic route generation.} (a)~A LIBERO demonstration, split into grasp, carry and place. A loop
(1--4) (b)~re-anchors the pieces in a new layout, (c)~bends the carry around an obstacle (MPLib check), and (d)~keeps
executions that pass all checks.}
\label{fig:datagen}
\vspace{-4.5mm}
\end{figure}

We therefore generate demonstrations that vary these factors (\cref{fig:datagen}). As in
MimicGen~\cite{mandlekar2023mimicgen,garrett2024skillmimicgen}, each demonstration re-anchors a human one to a new
layout: the grasp follows the object, the placement is solved for the new receptacle, and the human carry is stretched
between them. The receptacle is moved, and the start and clutter are re-sampled, so the goal must come from the scene
and the instruction~(iii). Obstacles from six families are placed on the nominal path, and each layout is generated
without and with them. For gate, pole and bar, a long object passes only by turning~(i, ii). Each scene also receives
alternative routes left of, right of or over each obstacle, or bowing in open scenes~(iv).

A detour bends the human carry around or over the obstacle, or changes when the object turns, and
MPLib~\cite{sulab2021mplib} checks it for the arm with the object attached. Every demonstration is then executed in
simulation~\cite{zhu2020robosuite,todorov2012mujoco} and kept only if it passes hard checks: success after release, a
single rigid grasp, no contact, and joint and speed limits. Its route is the held object's executed trajectory, so it is
feasible for the arm. No rule is task-specific. A task's generated routes differ by 23--31\,cm, against
2--4\,cm in LIBERO.

From LIBERO's 4,605 human demonstrations, the generator produces 16,049, plus 79,470 alternative routes. Each record
pairs the inputs $\cond=(\scene,\objcloud,\startpose,\instr)$ with one route of the held object, and the records are split
by layout into training and validation, so a layout's variants without and with obstacles fall in the same set.

\subsection{Hierarchical planning with a frozen VLA}
\label{sec:steering}
\label{sec:select}
\label{sec:vla_to_cairn}
\label{sec:cairn_to_vla}

\method and \pizero guide each other through the motion of the held object: the VLA's intent shapes the route, and the
route guides the VLA's actions. The VLA's conditional distribution
$p_\phi(\act\mid\obs)$~\cite{physicalintelligence2025pi05} takes images, proprioception and an instruction as input. Its
$H$-action flow state $\mathbf X_s$ evolves from noise at $s=0$ to the clean chunk $\mathbf X=\mathbf X_1$ at $s=1$, and decodes to
actions $\act=\operatorname{dec}_\pi(\mathbf X)$. A differentiable map turns a decoded
chunk into held-object poses,
\begingroup\setlength{\abovedisplayskip}{3pt plus 1pt minus 1pt}\setlength{\belowdisplayskip}{3pt plus 1pt minus 1pt}%
\begin{equation}
 \Phi(\act)_h=\Tee_h(\act)\,G_A.
 \label{eq:phimap}
\end{equation}\endgroup
Here $h$ indexes actions, $\Tee_h$ is the predicted end-effector pose after action $h$, and $G_A=(\Tee_A)^{-1}A$ is the
rigid grasp transform measured once at the grasp.

Their distributions $p_\theta^Y$ and $p_\phi^X$ define the joint model
\begingroup\setlength{\abovedisplayskip}{3pt plus 1pt minus 1pt}\setlength{\belowdisplayskip}{3pt plus 1pt minus 1pt}%
\begin{equation}
 q(\mathbf Y,\mathbf X)\propto p_\theta^Y(\mathbf Y\mid\cond)\,p_\phi^X(\mathbf X\mid\obs)\,e^{-\mathcal E(\mathbf Y,\mathbf X)}.
 \label{eq:joint}
\end{equation}\endgroup
The energy compares the route from \method, $\xi=\operatorname{dec}_A(\mathbf Y)$, with the one from \pizero,
$\xi_\pi = \Phi(\operatorname{dec}_\pi(\mathbf X))$. It motivates guiding each model by
the other's output, held fixed.

At the grasp, we record $\cond$ once. An unguided sample $\act^{\pi}\sim p_\phi(\act\mid\obs)$ describes how the VLA would
continue from the current state. It guides planning and is not executed. Its predicted object positions guide \method
through
\begingroup\setlength{\abovedisplayskip}{3pt plus 1pt minus 1pt}\setlength{\belowdisplayskip}{3pt plus 1pt minus 1pt}%
\begin{equation}
 E_{\mathrm C}(P)=\tfrac12\|BP-\Phi_{\text{pos}}(\act^{\pi})\|^2,
 \label{eq:intent}
\end{equation}\endgroup
where $P$ contains the clean route positions from \method and $\Phi_{\text{pos}}(\act^{\pi})$ the intent's positions
from \pizero, both centred at $\mathbf p_A$. The matrix $B$ interpolates route segments matched by translational arc
length, so the agreement leaves execution speed to the VLA. We draw $M$ routes, apply diagonally preconditioned
corrections to their clean estimates, and re-noise them with the same noise estimate. Routes that bring the held object
too close to the scene are rejected, and the medoid of the accepted routes becomes the plan. The supplement gives both
steps.

A second VLA solve then follows the selected route. At each flow step, the VLA velocity field $v_\phi$ gives the clean
estimate $\widehat{\mathbf X}_s=\mathbf X_s+(1-s)v_\phi(\mathbf X_s,s,\obs)$, whose decoded actions
$\acthat_s=\operatorname{dec}_\pi(\widehat{\mathbf X}_s)$ give the predicted object poses $\Phi(\acthat_s)$.
We match the next waypoint to one of these poses by position and orientation error, holding the match
fixed for the correction. This selects a predicted arrival without prescribing when the VLA must reach the waypoint.

Let $\mathbf r$ be the predicted position minus the waypoint and $E_{\mathrm{pos}}=\tfrac12\|\mathbf r\|^2$ the position
energy. Its gradient is $J^\top\mathbf r$, where
$J=\partial\mathbf r/\partial \operatorname{vec}(\widehat{\mathbf X}_s) \in\mathbb R^{3\times6H}$ contains the
derivatives of action decoding and $\Phi$. Under the local model
$\mathbf r(\widehat{\mathbf X}_s+\Delta) = \mathbf r(\widehat{\mathbf X}_s)+J\operatorname{vec}(\Delta)$, for a
correction $\Delta$ to the clean estimate, a gradient step moves the object by $-\eta JJ^\top\mathbf r$, so one gain
reduces errors unevenly across directions. We instead seek a small action correction that cancels a controlled fraction
$\kappa$ of the residual. Minimising
$\tfrac12\|J\operatorname{vec}(\Delta)+\kappa\mathbf r\|^2+\tfrac\varepsilon2\|\Delta\|_F^2$ over $\Delta$ yields the
damped Gauss--Newton step
\begingroup\setlength{\abovedisplayskip}{3pt plus 1pt minus 1pt}\setlength{\belowdisplayskip}{3pt plus 1pt minus 1pt}%
\begin{equation}
 \operatorname{vec}(\Delta_s^{\mathrm{pos}})
 =-\kappa J^\top(JJ^\top+\varepsilon I_3)^{-1}\mathbf r .
 \label{eq:gn}
\end{equation}\endgroup
The residual is three-dimensional, so one $3\times3$ solve corrects all contributing action coordinates, avoiding
repeated sequential guidance updates. Rotation uses a soft-matched geodesic energy, spreading its correction over
plausible arrivals. A phase-dependent gripper energy encourages closing during the carry and opening at release, and the
supplement gives both energies. Combining the three corrections,
$\Delta_s = \Delta_s^{\mathrm{pos}}+\Delta_s^{\mathrm{rot}}+\Delta_s^{\mathrm{grip}}$, the guided step to the next flow time $s'$ is
\begingroup\setlength{\abovedisplayskip}{3pt plus 1pt minus 1pt}\setlength{\belowdisplayskip}{3pt plus 1pt minus 1pt}%
\begin{equation}
 \mathbf X_{s'}=\mathbf X_s+(s'-s)\,v_\phi(\mathbf X_s,s,\obs)+\Delta_s.
 \label{eq:guided_vla_flow}
\end{equation}\endgroup
The correction retains derivatives through decoding and $\Phi$, including the position Jacobian $J$ above. Transport
through the VLA clean estimate would additionally involve
$\mathcal J_s^\pi=I+(1-s)\partial v_\phi/\partial\mathbf X_s$. We approximate $\mathcal J_s^\pi$ by the identity and
apply the correction directly to $\mathbf X_s$. This avoids a backward pass through the VLA and the need to recompute
its intermediate activations at each flow step. FlowChef~\cite{patel2025flowchef} uses a related direct clean-estimate
correction.

\pizero executes a short prefix of each chunk and observes again. \method refines the route less often on a separate
worker, and the VLA follows its route until a replacement arrives.
\section{Experiments}
\label{sec:exp}

We ask four questions. Does the prior propose good routes in unseen layouts (\cref{sec:exp_prior})? Does
steering with it make a frozen VLA succeed, compared with other steering and with fine-tuning
(\cref{sec:exp_steer})? Does it hold beyond our benchmark (\cref{sec:exp_beyond})? Which of its parts
matter (\cref{sec:exp_ablation})?

To test on unseen layouts, we run the generator of \cref{sec:data} again and draw 1,941 scenes, \heldout, that share
no layout or initial state with the training and validation data. They cover 27 tasks from the four LIBERO suites
\pizero was trained on, and about half contain an obstacle. Each scene is run with three seeds, 5,823 episodes in all,
and the ablations use one. Episodes start with the object lifted in the gripper. An episode succeeds once the held object rests in or on the
instructed receptacle and the gripper has released it, and fails if the robot or the object touches an obstacle
first. All methods run on the same episodes, and those that follow our prior use the same prior and seeds.

The prior (4.35M parameters) draws 16 routes of 20 waypoints in 10 Euler steps, and a route is clear if the held object stays at least
1.5\,cm from the scene. The VLA's intent pulls these routes at the grasp and then reweights them, and the plan
switches to another route once it is 3 times as probable. Once the routes separate, the plan is re-sampled every 4 chunks
from half noise. Steering corrects each VLA action chunk ($H{=}10$ steps, 5 executed) with \cref{eq:gn}, using
$\kappa{=}0.7$ and $\varepsilon{=}10^{-4}$\,m$^2$. Other settings are in the supplement.
\subsection{Routes on held-out scenes}
\label{sec:exp_prior}
\begin{table}[t]
\centering
\caption{\textbf{Routes on \heldout} at the grasp state. \emph{$\geq$1 valid}: episodes with a route keeping
the held object 1.5\,cm from the scene. \emph{Goal reached}: the selected route ends in the instructed receptacle
(turn: gate, pole or bar). \emph{Routes}: distinct routes among 16 samples.
\emph{Time}: per grasp, one B200 GPU (RRT: one CPU core).}
\label{tab:routes}
\vspace{-2mm}
\footnotesize
\setlength{\tabcolsep}{1pt}
\begin{tabular*}{\columnwidth}{@{\extracolsep{\fill}}l ccccc@{}}
\toprule
\multirow{2}{*}{Route source} & \multirow{2}{*}{$\geq$1 valid} & \multicolumn{2}{c}{Goal reached} &
\multirow{2}{*}{Routes} & \multirow{2}{*}{Time} \\
\cmidrule(lr){3-4}
 & & all & turn & & \\
\midrule
\method prior & \best{100.0\%} & \best{97.3\%} & \best{97.1\%} & \best{3.4} & \best{0.37\,s} \\
RRT~\cite{kuffner2000rrtconnect} & 88.4\% & 90.3\% & 91.6\% & 1.0 & 45.5\,s \\
3D HAMSTER~\cite{hwang2026hamster3d} & 37.5\% & 37.7\% & 48.5\% & 1.0 & 3.7\,s \\
\midrule
\quad LIBERO data only & 85.6\% & 16.3\% & 12.0\% & 1.0 & 0.36\,s \\
\quad 3DFA network~\cite{gkanatsios2025flowmatchactor} & 99.9\% & 95.5\% & 94.2\% & 3.5 & 0.29\,s \\
\quad single-route data & 98.6\% & 98.6\% & 99.3\% & 1.0 & 0.39\,s \\
\bottomrule
\end{tabular*}
\vspace{-2mm}
\end{table}

\cref{fig:routes} shows routes the prior proposes for held-out scenes: it tilts, turns or lifts the held object as
the receptacle requires, and passes an obstacle on either side.
\cref{tab:routes} scores the routes drawn at the grasp in every \heldout episode against the requirements of
\cref{sec:planner}: a clear route~(ii), a goal in the instructed receptacle~(iii), and several
distinct routes~(iv). The object's rotation~(i) is tested in closed loop (\cref{sec:exp_ablation}). The prior has a
clear route in all but one of the 5,823 episodes and reaches the receptacle in 97.3\%. RRT-Connect~\cite{kuffner2000rrtconnect,sulab2021mplib},
given the prior's goal and the same selection, finds a clear route in 88.4\% and is
slow: 45.5\,s per grasp on a CPU core, against 0.37\,s for the prior on a GPU. The released
3D HAMSTER planner~\cite{hwang2026hamster3d} reaches the receptacle in 37.7\% of episodes, and its path is clear in
37.5\%. Of the prior's ingredients, the data matters most: trained on LIBERO's demonstrations alone,
the same network reaches the receptacle in 16.3\%. Trained without alternative routes, it reaches the
receptacle slightly more often (98.6\%) but proposes a single route, with none clear in 1.4\%. With
3DFA's network~\cite{gkanatsios2025flowmatchactor}, it misses 1.6$\times$ as many goals, and twice as many where the
object must turn.

\subsection{Closed-loop steering}
\label{sec:exp_steer}

Each baseline changes one thing relative to \method: what \pizero is steered toward, or its weights.
OmniGuide~\cite{song2026omniguide} steers the denoising of \pizero with local energies on the gripper's predicted
positions: a repeller from the scene's point cloud and, optionally, an attractor to a target point chosen by a VLM,
which we replace with the receptacle's true centre. 3D HAMSTER~\cite{hwang2026hamster3d} is a hierarchical VLA whose
planner predicts a 3D gripper path (positions only, no orientation) from an image and the instruction. We
steer \pizero along this path with \method's steering step, so only the route differs. Finally, we fine-tune \pizero on the prior's
training scenes together with LIBERO's own demonstrations, which gives it the same layouts with actions. Baselines are tuned with equal budgets on a
development set.
\begin{table}[t]
\centering
\caption{\textbf{Steering a frozen VLA on \heldout.} \emph{Success}: released in the instructed receptacle with no
robot or object contact with an obstacle; \emph{detour}: wall, pillar or block; \emph{turn}: gate, pole or bar.
\emph{No contact}: obstacle-scene episodes. \emph{LIBERO}: the standard benchmark. \emph{Time}: per action chunk on one
B200.}
\label{tab:main}
\vspace{-2mm}
\footnotesize
\setlength{\tabcolsep}{1pt}
\begin{tabular*}{\columnwidth}{@{\extracolsep{\fill}}l ccc ccc@{}}
\toprule
 & \multicolumn{3}{c}{Success} & \multirow{2}{*}{No contact} &
\multirow{2}{*}{LIBERO} & \multirow{2}{*}{Time} \\
\cmidrule(lr){2-4}
 & all & detour & turn & & & \\
\midrule
\multicolumn{7}{@{}l}{\emph{frozen \pizero}} \\
\pizero alone & 24.5\% & 17.3\% & 13.4\% & 60.8\% & 96.5\% & \best{128\,ms} \\
OmniGuide~\cite{song2026omniguide} & 24.8\% & 18.0\% & 12.7\% & 63.6\% & -- & 243\,ms \\
OmniGuide + target & 50.9\% & 43.2\% & 30.4\% & 70.1\% & -- & 183\,ms \\
3D HAMSTER~\cite{hwang2026hamster3d} & 29.2\% & 8.8\% & 5.2\% & 24.6\% & -- & 159\,ms \\
\method & \best{95.1\%} & \best{92.4\%} & \best{88.8\%} & \best{92.9\%} & \best{97.0\%} & 190\,ms \\
\midrule
\multicolumn{7}{@{}l}{\emph{\pizero fine-tuned}} \\
\pizero alone & 91.3\% & 85.8\% & 78.2\% & 85.3\% & 70.5\% & \best{127\,ms} \\
\method & \best{96.1\%} & \best{93.8\%} & \best{89.9\%} & \best{93.8\%} & \best{70.9\%} & 187\,ms \\
\bottomrule
\end{tabular*}
\vspace{-2mm}
\end{table}  %

What \pizero lacks is the route, not the skill: it solves 96.5\% of standard LIBERO, yet succeeds in only 24.5\% of
\heldout (\cref{tab:main}). It carries the object to the wrong place in open scenes
(its failures end a median 45\,cm from the receptacle) and straight into obstacles in 39\% of obstacle
episodes. \method supplies the missing route and raises success to 95.1\%. It helps most where the object must
turn: from 13.4 to 88.8\%.

Other targets do not replace the route. Local fields leave \pizero's route as it was: OmniGuide's repeller pushes the
gripper away from nearby geometry but does not move the goal, and it scores the gripper rather than the held object, so
it does not turn the object (24.8\% overall, 12.7\% where the object must turn). An attractor to the
receptacle, even at its true centre, reaches only 50.9\%, because a straight pull stalls behind obstacles, the local
minima its authors note~\cite{song2026omniguide}. Along 3D HAMSTER's path, the same steering adds little overall (29.2 vs.\ 24.5\%) and hurts around obstacles
(8.8 vs.\ 17.3\% in detour scenes). Without orientation, its paths cannot turn a long object through a gate (5.2 vs.\ 13.4\%).

Changing the weights is not a substitute for a route, but it composes with one. Fine-tuning on the prior's scenes
teaches \pizero the new layouts: 91.3\% success. But it needs robot
actions for every layout, still does worse than \method around obstacles (85.8 and 78.2 vs.\ 92.4 and 88.8\% in
detour and turn scenes), and loses much of what \pizero knew: success on standard LIBERO drops from 96.5 to 70.5\%,
while \method keeps 97.0\%. Steering the fine-tuned model along \method's route gives the highest success
(96.1\%).

\subsection{Beyond the benchmark}
\label{sec:exp_beyond}
\begin{table}[t]
\centering
\caption{\textbf{Beyond the benchmark:} success of frozen \pizero, alone and steered. \emph{LIBERO-PRO}: position
axis, on all episodes, on the 27 of its 40 tasks within our scope, and on those where \pizero lifts the instructed
object, which \method steers. \emph{Real robot}: a Franka FR3 places a
cube in a bowl, behind a box, and on a cabinet top, with the cube released on the target and the box untouched.}
\label{tab:shift}
\label{tab:real}
\vspace{-2mm}
\footnotesize
\setlength{\tabcolsep}{9pt}
\newcommand{\bsub}{\hspace{1em}}
\begin{tabular*}{\columnwidth}{@{\extracolsep{\fill}}l cc@{}}
\toprule
 & \pizero alone & \pizero + \method \\
\midrule
\multicolumn{3}{@{}l}{\emph{LIBERO-PRO, position axis}} \\
\bsub all episodes & 27.3\% & \best{38.0\%} \\
\bsub single-carry tasks & 32.2\% & \best{48.0\%} \\
\bsub + correct grasp & 54.4\% & \best{81.1\%} \\
\midrule
\multicolumn{3}{@{}l}{\emph{Real robot, \pizero-DROID}} \\
\bsub bowl & \best{10/10} & \best{10/10} \\
\bsub bowl behind a box & 0/10 & \best{9/10} \\
\bsub cabinet top & 0/10 & \best{8/10} \\
\bottomrule
\end{tabular*}
\vspace{-2mm}
\end{table}

\paragraph{Third-party layout shift.}
LIBERO-PRO~\cite{zhou2025liberopro} perturbs LIBERO's own tasks. Its position axis swaps the initial positions of
objects, a shift that neither our generator nor the prior's training scenes were built for. Episodes start before the
grasp and use LIBERO's own success check and step limit. We run its 40 tasks, 50 episodes each. \pizero succeeds in 27.3\% of episodes, and \method raises this to 38.0\%
(\cref{tab:shift}). The gain is bounded by what steering can change. Steering starts after the grasp, so episodes in
which \pizero picks the wrong object, or none, stay failures, and 13 of the 40 tasks are not single carries (two
objects, drawers, the stove) and run unsteered. Where \method applies, after a correct grasp on a single-carry task,
success rises from 54.4\% to 81.1\%.

\begin{figure}[t]
\centering
\newcommand{\realpanel}[3]{%
\tikz[inner sep=0pt]{\node (p) {\includegraphics[width=\dimexpr0.487\columnwidth\relax]{figures/real_#1.jpg}};
\draw[#2, line width=0.9pt] (p.south west) rectangle (p.north east);
\node[anchor=north west, fill=#2, text=white, inner sep=1.6pt, font=\fontsize{6}{6}\selectfont] at (p.north west) {#3};}}
\footnotesize
\setlength{\tabcolsep}{0pt}
\begin{tabular}{@{}c@{\hspace{3pt}}c@{}}
\pizero alone & \pizero + \method \\[1pt]
\multicolumn{2}{c}{\textbf{``put the cube in the bowl''}, box in the way} \\[1pt]
\realpanel{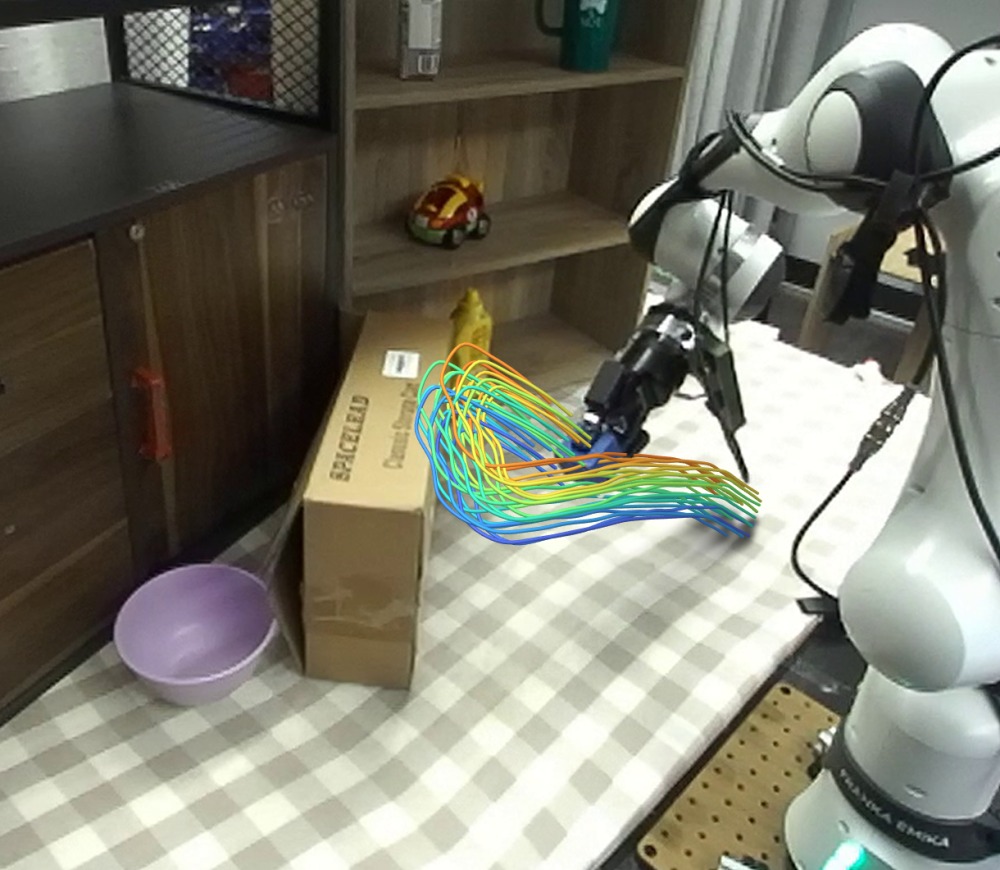}{red!72!black}{\xmark} & \realpanel{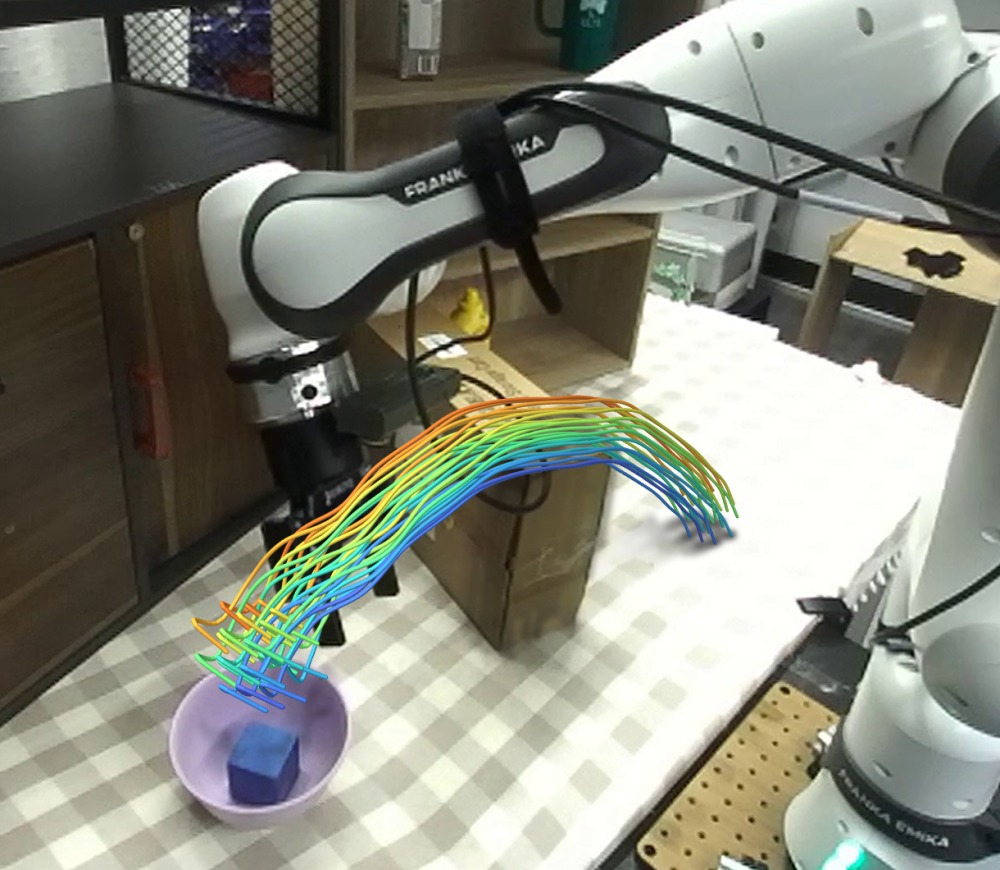}{green!48!black}{\cmark} \\[3pt]
\multicolumn{2}{c}{\textbf{``put the cube on top of the cabinet''}} \\[1pt]
\realpanel{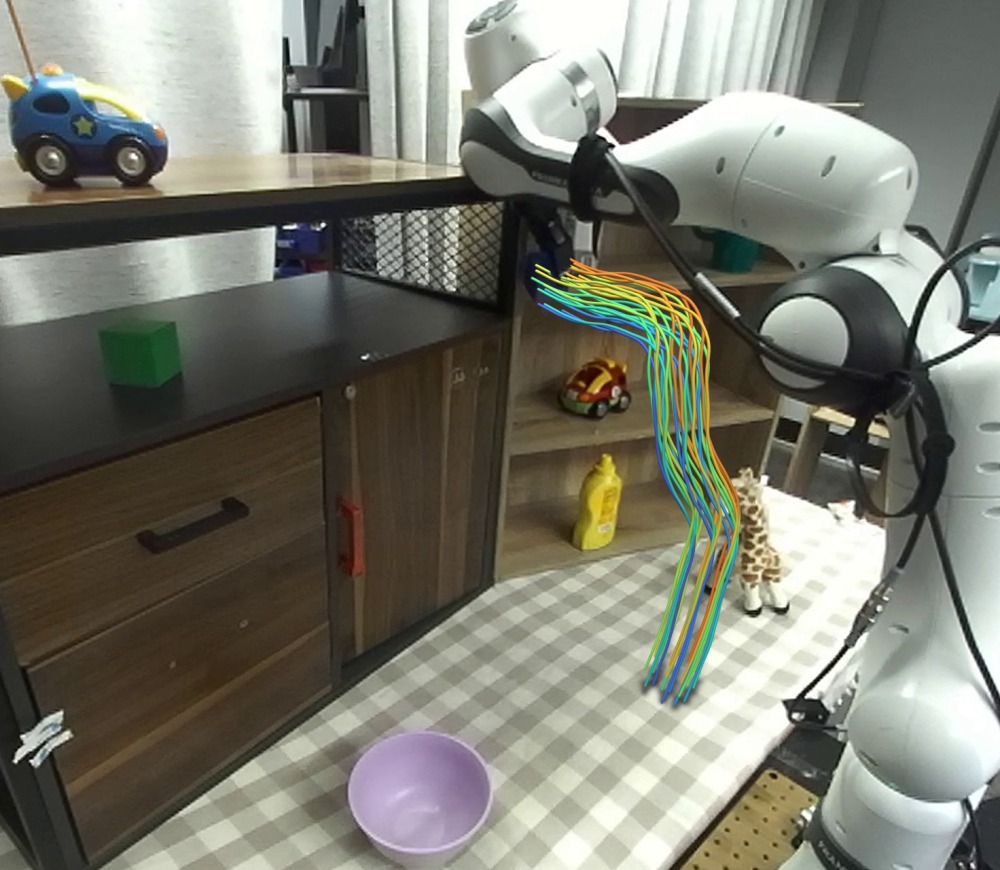}{red!72!black}{\xmark} & \realpanel{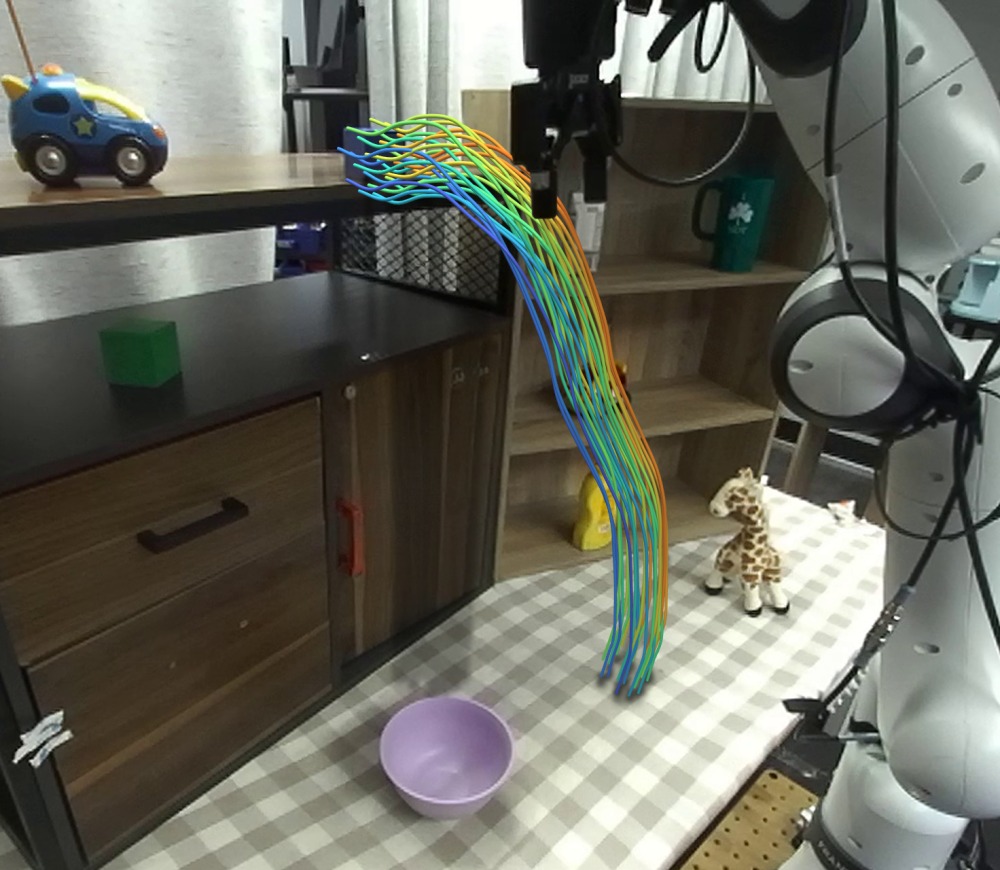}{green!48!black}{\cmark} \\
\end{tabular}
\caption{\textbf{Real robot:} 3D tracks of points on the cube, up to the frame shown. \pizero alone runs
the cube into the box and stops at the cabinet's edge; with \method it clears the box and reaches the top.}
\label{fig:real}
\vspace{-4mm}
\end{figure}

\paragraph{Real robot.}
Our platform (\cref{fig:real}) is a DROID
setup~\cite{khazatsky2024droid} with a Franka FR3 arm, driven by the released \pizero-DROID~\cite{physicalintelligence2025pi05},
which stays frozen. We fine-tune the prior on demonstrations that our generator produces in a replica of the lab, a
scan of the real scene with rendered receptacles and obstacles, without real demonstrations. When the object is grasped, the scene and object clouds come from the cameras' depth, and the object's pose
from the arm's kinematics and the grasp measured once, as the supplement details. The prior samples 32 routes, and \method steers toward the medoid of those that
clear the scene and that the arm can reach, if any. We run
10 trials per task, with and without \method, on three tasks with a cube (\cref{tab:real}): placing it in a
bowl, in the same bowl with a box in the way, and on top of a cabinet. A trial succeeds when the cube is released in
the bowl or on the cabinet's top board and the box has not been touched. With a free path, \pizero alone succeeds in
10/10 trials and \method leaves this unchanged. With the box in the way it succeeds in 0/10: it runs
the cube into the box or pushes the box aside on its way to the bowl (\cref{fig:real}). The cabinet's top board lies
near the limit of the arm's reach, and \pizero alone fails to bring the arm out that far (0/10): it stops at
the edge of the board, still holding the cube. With \method it succeeds in 9/10 and 8/10: the route
passes over the box, and at the cabinet it rises above the top board and leads the arm out to the target.
The failures are a box graze and a release at the board's edge.

\subsection{Ablations}
\label{sec:exp_ablation}
\begin{figure}[t]
\centering
\includegraphics[width=\linewidth,height=0.9\linewidth]{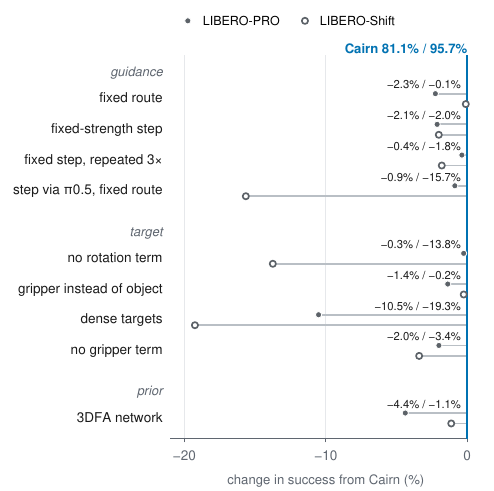}
\vspace{-4mm}
\caption{\textbf{Ablations:} change in success from \method on LIBERO-PRO (filled: the 799 steered episodes)
and \heldout (open: one seed per scene). Step via \pizero: RTC~\cite{black2025rtc}, on the fixed route.}
\label{fig:ablations}
\vspace{-5.5mm}
\end{figure}

\cref{fig:ablations} changes one part of \method at a time. Following the route fixed at the grasp, without the
coupling, costs 2.3 points on LIBERO-PRO (43 vs.\ 26 episodes solved by only one) and ties on \heldout (95.6
vs.\ 95.7\%). Where \pizero's intent makes \method follow a different route, in about a fifth of \heldout scenes, the
hand moves with 12\% less jerk, the early corrections are 16\% smaller, and the arm and the object keep 0.8\,cm more
clearance. \heldout leaves little room for success to change. Most of \method's failures there are not the route's to
fix, such as the arm touching an obstacle while the object follows the route. Similarly, on
LIBERO-PRO, where success has room, our changes to the 3DFA network (the Point Transformer V3 scene encoder, the
held object's own branch and the local attention) are worth 4.4 points (62 vs.\ 27 episodes solved by only
one).

The normalised step of \cref{eq:gn} covers a fixed fraction of the distance to the waypoint in one solve. A gradient
step of fixed strength costs 2.1 and 2.0 points, and on \heldout the carried object touches obstacles almost four times
as often (38 vs.\ 10 episodes). Repeating it three times per denoising step, to approach the normalised step by
iteration, recovers most of the loss on LIBERO-PRO but not on \heldout ($-1.8$), and takes 18\% longer per chunk. RTC,
on the route fixed at the grasp, pulls the correction back through \pizero's network and ties each waypoint to a time
step of the chunk. It is 35\% slower than \method and loses 15.7 points on \heldout: through the network the correction
shrinks, and the time steps impose the route's pace on \pizero.

The prior is a high-level planner, and it helps most when used as one. Placing targets densely along the route, as
low-level control would, costs the most on both benchmarks (10.5 and 19.3 points). A route has no schedule, so every
extra target is a demand on \pizero's pace, and on \heldout the object is still being carried at the time limit in 365
of the 457 failures. Pace is \pizero's job, and sparse waypoints leave it to \pizero. Rotation and the object frame
matter where the object turns. On \heldout, success drops by 13.8 points without the rotation term, and steering the
gripper instead of the object costs 3.1 points in the turn scenes, shown in the supplement. On LIBERO-PRO, whose
swaps need no turns, both stay within 1.4 points. Without the gripper term \pizero lets go early, and success drops by
2.0 and 3.4 points.
\section{Conclusion and future work}
\label{sec:conclusion}
We presented \method, which steers a frozen VLA with a learned prior over the 3D route of the object in its gripper.
The prior and the VLA form the two levels of one planner. The prior proposes $\SE$ routes from the scene, the held
object and the instruction, and the VLA's intent selects and refines them. The chosen route then pulls the VLA's
denoising toward its next waypoint, while the VLA still grasps and sets its own pace. Because the route belongs
to the object, the prior needs no robot actions and turns long objects through gaps, and it transfers from simulation
to a real robot. Because the VLA is never retrained, it keeps its skills. In the future, the prior could also learn from
videos of people moving objects, toward a foundation model of object motion paired with generalist VLAs.
\par
{
    \small
    \bibliographystyle{ieeenat_fullname}
    \bibliography{main}
}

\end{document}